%% file: cinecue.tex
\documentclass{article}
\usepackage[preprint]{neurips_2026}
\usepackage[T1]{fontenc}
\usepackage[utf8]{inputenc}
\usepackage{microtype,amsmath,amssymb,booktabs,array,graphicx,xcolor}
\usepackage{tikz,pgfplots}
\pgfplotsset{compat=1.18}
\definecolor{ccReport}{HTML}{173F5F}
\definecolor{ccRateMatched}{HTML}{61A6AB}
\definecolor{ccCrossFitted}{HTML}{D7A86E}
\usetikzlibrary{positioning,arrows.meta,fit,backgrounds}
\usepackage[colorlinks=true,allcolors=blue]{hyperref}
\usepackage{url,placeins}
\newcolumntype{L}[1]{>{\raggedright\arraybackslash}p{#1}}
\newcommand{\reportdisplayspacing}{%
  \setlength{\abovedisplayskip}{7pt plus 1pt minus 1pt}%
  \setlength{\belowdisplayskip}{7pt plus 1pt minus 1pt}%
  \setlength{\abovedisplayshortskip}{7pt plus 1pt minus 1pt}%
  \setlength{\belowdisplayshortskip}{7pt plus 1pt minus 1pt}%
}
\AddToHook{cmd/normalsize/after}{\reportdisplayspacing}
\AddToHook{cmd/small/after}{\reportdisplayspacing}
\input{numbers}
\title{Closing the Perception--Report Gap in Chest CT Reporting through Autoregressive Finding Decisions}
\author{%
  Seungho Lee$^{1}$\thanks{Corresponding author: \href{mailto:aiden@lafit.dev}{\texttt{aiden@lafit.dev}}.}
  \qquad Joseph David Janizek$^{2}$ \qquad Eun Kyoung Hong$^{2}$ \\[4pt]
  \normalfont $^{1}$Lafit Inc. \qquad $^{2}$Stanford University
}
\hypersetup{pdfauthor={Seungho Lee, Joseph David Janizek, Eun Kyoung Hong}}
\begin{document}
\maketitle

\begin{abstract}\label{sec:abstract}
Chest CT report generators can omit abnormalities that remain recoverable from their visual representations. We close this perception--report gap by training the report decoder to make explicit finding decisions before composing the report. In \emph{decision-first generation}, a video language model answers yes or no for each finding, then continues into a descriptive state and the report, conditioned on its own answers. Finding-specific positive-answer weighting aligns these decisions with the objective of recovering abnormalities. Trained by supervised fine-tuning with CT-RATE as its sole task-specific CT corpus, the full-scale generator achieves \textbf{\N{mainF} report-label macro-F1}, establishing state-of-the-art single-model performance on the corrected CT-RATE protocol without additional preference or reinforcement learning. Matched component controls identify the coupling of explicit decisions and positive-answer weighting as the source of the gain. The full-scale model recovers findings in its reports at the level of a supervised visual readout used for analysis. Editing individual decisions redirects the corresponding report content. Complementary interventions on finding presence and reference templates distinguish gains in finding recovery from gains in reference-template agreement. Together, these results identify the decoder's content decisions as a practical target for learning stronger CT report generators. Code and model weights will be released.
\end{abstract}

\section{Introduction}\label{sec:introduction}
When a generated chest CT report omits an abnormality, where did the information disappear? The scan may not have been encoded well enough to identify the finding, or the finding may have been lost when visual information was turned into words. Hidden-state analyses show that supervised readouts recover information missed by generated answers \citep{fu2025hiddenplainsight}. SelfCue identifies a related gap in CT reporting and reduces it through probe-conditioned decoding and distillation \citep{selfcue}. Recovering a finding from visual representations and expressing it in a report are therefore distinct learning problems.

Free-text report supervision leaves the decision to affirm or omit a finding implicit in the likelihood of its description. We hypothesize that supervising these content decisions explicitly can close the perception--report gap. We train the report decoder to generate a yes/no answer for every finding in a fixed vocabulary, and then to continue the same autoregressive sequence into a descriptive state and the report. The answers provide explicit context for the ensuing text, and finding-specific positive-answer weights determine how readily findings are affirmed. We call this ordered generation process \emph{decision-first generation}. The design couples the form of a content decision with the objective used to learn it.

Implemented in a general video language model, this method establishes state-of-the-art single-model report-label performance on CT-RATE with supervised fine-tuning and CT-RATE-only task training. We explain this result through matched generation targets and component controls, testing how the decision format and its learning objective determine which recoverable findings reach the report.

The distinction between selecting findings and expressing them also matters for evaluation: do higher scores reflect better finding recovery or closer adherence to a reference's reporting conventions? We address this question by separately supplying reference findings and reference-template information. The former strongly improves finding recovery; the latter substantially raises text-agreement scores with little change in recovered abnormalities. These interventions distinguish gains in finding recovery from agreement with a reference's reporting pattern.

Our contributions are threefold:
\begin{itemize}\setlength\itemsep{2pt}
\item We introduce decision-first generation, in which one autoregressive model generates an exhaustive sequence of yes/no finding decisions and develops them into a descriptive state and the report.
\item We show that decision-first generation closes the measured perception--report gap, that its gain requires combining explicit decisions with positive-answer weighting in matched component controls, and that editing these decisions redirects report content.
\item We show that reference-template agreement can raise text-agreement scores with little change in abnormality recovery, motivating separate evaluation of these dimensions.
\end{itemize}

\section{Autoregressive Finding Decisions for CT Reporting}\label{sec:method}
\subsection{Finding decisions within the report sequence}
Let $x$ denote a CT volume, $z=f_\theta(x)$ its visual tokens, and $r$ the report. A conventional generator directly models $p_\theta(r\mid z)$. We introduce a decision sequence $d=(d_1,\ldots,d_L)$ for a fixed vocabulary of $L$ findings, followed by a descriptive state $s$ and the report:
\begin{equation}\label{eq:factorization}
p_\theta(d,s,r\mid z)=p_\theta(d\mid z)\,p_\theta(s\mid z,d)\,p_\theta(r\mid z,d,s).
\end{equation}
The report decoder generates each $d_\ell$ as an explicit yes/no answer, conditioned on the CT and preceding tokens. The descriptive state records finding names, assertion status, anatomical region, and short report-linked evidence phrases. The report consists of Findings and Impression. Both the state and the report attend to the visual tokens and all preceding generated tokens.

For CT-RATE, the \emph{decision prefix} answers all 18 report-derived findings in a fixed order. This exhaustive prefix gives every finding a supervised decision, including findings assigned a negative reference label. Each answer occupies one token, providing a specific location at which training can govern whether the decoder affirms the finding. The descriptive state then connects these presence decisions to the anatomical and descriptive detail needed for the report.

\subsection{Cost-sensitive supervision at the decision tokens}\label{sec:cost}
Making an answer explicit also exposes its decision boundary. Under binary cross-entropy, a greedy yes/no decision changes at probability one half. The threshold that maximizes finding F1 can differ substantially, particularly for less prevalent findings \citep{thresholdf1}. We encode a finding-specific boundary through the loss on positive answers.

For a binary answer with underlying positive probability $\eta$, let $q$ be the model's probability of yes conditional on choosing yes or no. Let $c_\ell>0$ denote the relative cost of a missed positive finding. Using this cost as the positive-answer weight gives the expected loss
\begin{equation}
\mathcal L_\ell(q)=-c_\ell\eta\log q-(1-\eta)\log(1-q),\qquad
q^*=\frac{c_\ell\eta}{c_\ell\eta+1-\eta}.
\end{equation}
Thus $q^*>1/2$ when $\eta>1/(1+c_\ell)$. To target a decision threshold $\tau_\ell$, we choose
\begin{equation}\label{eq:cost}
c_\ell=\frac{1-\tau_\ell}{\tau_\ell}.
\end{equation}
This classical cost-sensitive rule \citep{elkan2001costsensitive} makes the risk-minimizing answer switch to yes when the underlying positive probability exceeds $\tau_\ell$. Applying it to the answer token makes the learned boundary part of the generated context: subsequent text is conditioned on a finding decision trained toward that operating point.

We estimate $\tau_\ell$ by maximizing finding-specific F1 on pooled, patient-disjoint out-of-fold predictions from supervised visual readouts fitted to frozen representations of a state-first checkpoint, using training patients only. This checkpoint was trained to generate a descriptive state followed by the report. The resulting costs are fixed before full-scale training. They multiply the positive-answer loss within the common state-section weighting; all other tokens retain the shared training objective. At inference, greedy decoding writes the decisions and continues directly into the state and report.

\subsection{CT representation, supervision, and inference}\label{sec:backbone}
\paragraph{From CT slices to visual tokens.}
We adapt the public Qwen3-VL-8B-Instruct model \citep{bai2025qwen3vl} to an ordered axial CT sequence (Figure~\ref{fig:method}). Starting from a stored broad-window grayscale volume, we sample across its full depth, resize the selected slices, and group consecutive slices into fixed-length clips. All clips enter one model input, preserving coverage of the volume. The vision encoder embeds adjacent-slice patches, and the visual merger reduces their spatial grid. Timestamp fields carry source-slice indices across clip boundaries, so the model receives axial order rather than elapsed video time. Training and inference use the same deterministic preprocessing, with no additional frame sampling by the processor.

\input{figures/method}

\paragraph{Training targets and objective.}
CT-RATE labels supervise the decision prefix, and the reference Findings and Impression supervise the report. GPT-5 mini \citep{openai2025gpt5mini} produces the state annotations offline from training reports and finding labels; schema and source-span checks keep the state annotations linked to the source report. All task-specific CT examinations and source reports come from CT-RATE. We fine-tune the vision encoder, visual merger, and language model together using weighted token cross-entropy. Section weights and report-linked evidence-span weights are shared across the controlled models; finding-specific positive-answer weights act on positive decision tokens. This preserves a common reporting objective while changing how finding selection is supervised.

\paragraph{Inference.}
Given one CT and a fixed instruction, the model generates its decision prefix, descriptive state, and final report in a single autoregressive sequence using greedy decoding. Each stage retains access to the CT and preceding generated text. The model supplies its own intermediate decisions; no reference findings, external classifier predictions, or template identifiers enter ordinary report generation. We extract Findings and Impression for evaluation. Input sizes, optimization settings, and the common generation cap are listed in Appendix~\ref{app:implementation}.

\section{Experimental Setup}\label{sec:setup}
\paragraph{Data and evaluation protocol.}
CT-RATE pairs non-contrast chest CT volumes with radiology reports and 18 report-derived abnormality labels \citep{hamamci2026ctrate}. Multiple reconstructions can share an examination and report. Our training set contains \N{trainVolumes} reconstructions from \N{trainStudies} examinations and \N{trainPatients} patients. We generate reports for all \N{historicalVolumes} validation volumes and score both this original population and the corrected \N{evalVolumes}-volume population used by NV-Reason-CT, which excludes non-chest volumes \citep{nvreasonct}. The benchmark table reports both populations. Tables~\ref{tab:readout} and~\ref{tab:factorial} use the corrected set of \N{evalVolumes} reconstructions from \N{evalPatients} patients; Table~\ref{tab:report-metrics} reports results on all \N{historicalVolumes} reconstructions.

\paragraph{Report evaluation.}
The primary endpoint is \emph{report-label macro-F1}: the official CT-RATE RadBERT classifier extracts 18 labels from generated Findings and Impression at threshold 0.5, and we average finding-level F1 against the reference labels. Intermediate decisions and states are excluded. We call a report label-positive for a finding when the classifier extracts that label, and label-negative otherwise. Macro precision and recall characterize the reporting trade-off, while CRG provides the benchmark's distribution-aware label score \citep{crg}. We also measure BLEU-4, ROUGE-L, METEOR, and RadGraph-XL complete F1, which compare wording and extracted entities and relations with the reference \citep{papineni2002bleu,lin2004rouge,banerjee2005meteor,delbrouck2024radgraphxl}. GREEN is an LLM-based report judge \citep{ostmeier2024green}. Lexical metrics score Findings and Impression; RadGraph-XL and GREEN score Findings. These text metrics and the paired content/template comparisons use all original validation volumes. Section~\ref{sec:metrics} further examines GREEN through a controlled template intervention.

\paragraph{Measuring the perception--report gap.}
For analysis, we train a supervised visual readout to predict the same findings directly from each generator's visual tokens. Learned finding queries attend over the tokens of the whole CT and produce one logit per finding. The readout is trained alongside each generator on detached tokens, with the same architecture and objective across conditions; it supplies neither backbone gradients nor report conditioning. This provides a common instrument for comparing recoverable finding information with the findings expressed in the report.

We examine two readout operating points. A \emph{rate-matched} readout predicts each finding in exactly as many volumes as the corresponding reports do, testing whether the same number of positive calls can be placed more accurately. A \emph{cross-fitted} readout uses finding-specific F1-maximizing thresholds fitted on four fifths of the validation patients and evaluated on the remaining fifth, repeated across five patient-disjoint folds. The first comparison holds reporting frequency fixed; the second allows thresholds to adapt. Readout AUROC additionally measures ranking independently of either threshold choice.

\paragraph{Controlled training and interventions.}
Three full-scale models share the public initialization, CT inputs, training data, update budget, and learning-rate schedule. They generate a report alone (direct-report), a descriptive state followed by the report (state-first), or finding decisions followed by a descriptive state and the report (decision-first, with positive-answer weighting). All are evaluated at update \N{selectedUpdate}. A separate four-condition component experiment crosses the presence of the decision prefix with finding-specific costs. These models start from one shared state-first checkpoint and train for 649 updates with identical batches and data order. Without a prefix, the costs act on positive finding-name tokens in descriptive states; with a prefix, they act on positive answers. This contrast tests the role of the decision location beyond simply emphasizing finding-related text.

We then intervene on the generated prefix and on the CT input to test the decision-to-report and image-to-decision connections. A template-conditioned variant, trained at full scale with a template identifier after its decision prefix, supports the separate evaluation analysis. Each intervention's supplied information and held-fixed context are defined with its results.

\paragraph{Statistical comparisons.}
We report 95\% patient-cluster bootstrap intervals, retaining each patient's evaluated reconstructions together. Differences between our models use paired patient resampling. External scores are taken from the cited evaluations, with their population and source identified. The component controls and interventions test predictions specified before their outputs were scored.

\section{Results}\label{sec:main-results}
\subsection{Benchmark performance and text agreement}\label{sec:benchmark}
Decision-first generation achieves the strongest reported single-model report-label performance on both CT-RATE validation protocols (Table~\ref{tab:benchmark}). This result is obtained with CT-RATE-only task training, supervised fine-tuning, and no external diagnostic cues at inference. Table~\ref{tab:benchmark} separates these requirements: AbSteering adds direct preference optimization (DPO), NV-Reason-CT uses additional CT cohorts and group relative policy optimization (GRPO), and DCP-PD supplies predictions from a separate classifier.

\begin{table}[t]
\centering\small
\caption{CT report generation: training and inference requirements alongside report-label performance. Training columns describe downstream CT-task adaptation, excluding the upstream training of released backbones. Pref./RL denotes additional preference or reinforcement learning; external cues are diagnostic predictions supplied to a report decoder by a separate model at inference. F1 is report-label macro-F1; -- denotes an unreported score.}
\label{tab:benchmark}
\setlength{\tabcolsep}{2.5pt}
\begin{tabular}{@{}lcccrrr@{}}
\toprule
& \multicolumn{2}{c}{CT-task training} & External & Original & \multicolumn{2}{c}{Corrected (\N{evalVolumes})} \\
\cmidrule(lr){2-3}\cmidrule(l){6-7}
Model & CT images & Pref./RL & cues & F1 (\N{historicalVolumes}) & F1 & CRG \\
\midrule
CT2Rep$^\dagger$ \citep{hamamci2024ct2rep} & CT-RATE & None & None & 0.1065 & -- & -- \\
CT-CHAT$^\dagger$ \citep{hamamci2026ctrate} & CT-RATE & None & n/s & 0.2166 & -- & -- \\
M3D$^\dagger$ \citep{bai2024m3d} & CT-RATE & None & None & 0.2674 & -- & -- \\
InternVL3-8B$^\dagger$ \citep{zhu2025internvl3} & CT-RATE & None & None & 0.3891 & -- & -- \\
\addlinespace[3pt]
AbSteering \citep{fang2026absteering} & CT-RATE & DPO & None & 0.4766 & -- & -- \\
NV-Reason-CT \citep{nvreasonct} & Mixed$^\ddagger$ & GRPO & None & -- & 0.5923 & 0.5138 \\
\addlinespace[3pt]
DCP-PD \citep{wang2026dcppd} & CT-RATE & None & AnyMC3D & 0.603 & -- & -- \\
\addlinespace[3pt]
\multicolumn{7}{@{}l}{\emph{Classifier-driven retrieval}}\\
ARC-CT team retrieval \citep{isik2026arcctchallenge} & CT-RATE & None & n/a & -- & 0.6176 & 0.5426 \\
\midrule
\multicolumn{7}{@{}l}{\emph{Our matched Qwen3-VL-8B models}}\\
Direct-report & CT-RATE & None & None & \N{qualityFullF-direct} & \N{benchmarkCorrected-direct} & \N{benchmarkCRG-direct} \\
State-first & CT-RATE & None & None & \N{qualityFullF-state} & \N{benchmarkCorrected-state} & \N{benchmarkCRG-state} \\
\textbf{Decision-first} & CT-RATE & None$^{*}$ & None & \textbf{\N{mainFullF}} & \textbf{\N{mainFPrecise}} & \textbf{\N{mainCRG}} \\
\bottomrule
\end{tabular}
\par\vspace{4pt}
\begin{minipage}{\linewidth}\footnotesize
$^\dagger$CT-RATE retraining by \citet{fang2026absteering}; their AbSteering row uses InternVL3-8B. Whether the evaluated CT-CHAT receives external cues is not specified (n/s). $^\ddagger$CT-RATE, an internal NIH cohort, and CancerVerse. DCP-PD uses the DCP-2 setting (AnyMC3D predictions); its \N{historicalVolumes}-volume denominator is inferred from the published class counts. M3D and DCP-PD transfer CT-pretrained backbones. The decision-first model's corrected-population F1 interval is \N{mainLow}--\N{mainHigh} (95\%, patient-cluster bootstrap). ARC-CT team retrieval uses one image classifier to select training reports, with no report decoder (n/a). The same team reports 0.6287 F1 and 0.5641 CRG with a classifier ensemble, consensus selection, and sentence editing.
\par $^{*}$The full-scale decision-first generator was trained by supervised fine-tuning; its 18 positive-answer costs were estimated from readouts of a separate checkpoint that had undergone self-critical training (Appendix~\ref{app:implementation}).
\end{minipage}
\end{table}

The matched models isolate the benefit of the decision-first training design. Relative to descriptive-state supervision, decision-first generation improves report-label macro-F1 by \N{fullGain} (\N{fullGainLow} to \N{fullGainHigh}) under the same architecture, CT inputs, and training budget. The following analyses explain how this gain arises and why the two training components interact.

\begin{table}[t]
\centering\small
\caption{Report-label and text-agreement scores of the three matched full-scale models on all \N{historicalVolumes} validation volumes. F1 is report-label macro-F1; RG is RadGraph-XL complete F1. All metrics use the same generated reports.}
\label{tab:report-metrics}
\begin{tabular}{@{}lrrrrr@{}}
\toprule
Model & F1 & BLEU-4 & ROUGE-L & METEOR & RG \\
\midrule
Direct-report & \N{qualityFullF-direct} & \N{qualityBleu-direct} & \N{qualityRouge-direct} & \N{qualityMeteor-direct} & \N{tmRG-reportOnly} \\
State-first & \N{qualityFullF-state} & \N{qualityBleu-state} & \N{qualityRouge-state} & \N{qualityMeteor-state} & \N{tmRG-noScreen} \\
Decision-first & \N{mainFullF} & \N{qualityBleu-main} & \N{qualityRouge-main} & \N{qualityMeteor-main} & \N{tmRG-main} \\
\bottomrule
\end{tabular}
\end{table}

The broader report metrics expose a different ordering (Table~\ref{tab:report-metrics}). The direct-report model obtains higher text-agreement scores despite recovering far fewer reference abnormalities. The state-first and decision-first models improve finding recovery without a corresponding increase in RadGraph-XL. This divergence motivates the controlled analysis of finding content and reference templates in Section~\ref{sec:metrics}.

\subsection{Closing the perception--report gap}\label{sec:readout}
Table~\ref{tab:readout} places the generated reports beside supervised readouts of the same models' visual tokens. The direct-report model has a strong visual readout but low report recall: useful finding information is recoverable without appearing in its text. Descriptive states improve report recovery substantially, and weighted finding decisions bring the report to the level of the cross-fitted readout. The progression connects the benchmark gain to how the decoder turns visual information into report content.

\begin{table}[t]
\centering\small
\caption{Finding recovery from generated reports and visual readouts on the corrected \N{evalVolumes} volumes. P and R are macro precision and recall. Rate-matched readouts make the same number of positive calls per finding as their reports; cross-fitted thresholds are fitted on other patients. AUROC measures the readout's ranking.}
\label{tab:readout}
\setlength{\tabcolsep}{5pt}
\begin{tabular}{@{}lrrrrrr@{}}
\toprule
& \multicolumn{3}{c}{Generated report} & \multicolumn{3}{c}{Visual readout} \\
\cmidrule(lr){2-4}\cmidrule(l){5-7}
Model & F1 & P & R & Rate-matched F1 & Cross-fitted F1 & AUROC \\
\midrule
Direct-report & \N{reportOnlyF} & \N{readP-direct} & \N{readR-direct} & \N{rateHead-reportOnly} & \N{rateCross-reportOnly} & \N{reportOnlyHeadAUROC} \\
State-first & \N{noScreenF} & \N{readP-state} & \N{readR-state} & \N{rateHead-noScreen} & \N{rateCross-noScreen} & \N{noScreenHeadAUROC} \\
Decision-first & \N{mainF} & \N{readP-main} & \N{readR-main} & \N{rateHead-main} & \N{rateCross-main} & \N{mainHeadAUROC} \\
\bottomrule
\end{tabular}
\end{table}

The rate-matched comparison distinguishes selection quality from simply stating more findings. For the direct-report model, rate matching retains its low reporting frequency, whereas cross-fitting also allows the readout to make more positive calls. Even when it makes the same number of positive calls, the state-first readout exceeds its report by \N{rateLoss-noScreen} (\N{rateLossLow-noScreen} to \N{rateLossHigh-noScreen}). The corresponding difference for decision-first generation is \N{rateLoss-main} (\N{rateLossLow-main} to \N{rateLossHigh-main}). The gap therefore closes at matched reporting frequency as well as under cross-fitted thresholds. Readout AUROC also improves, so the overall gain combines stronger visual representations with more effective use of those representations in the report.

A paired example makes the distinction concrete. For one CT, the direct-report readout recovers medical material, arterial and coronary wall calcification, and lung nodules, yet its report describes a normal examination. The decision-first model affirms and reports all four reference-labeled findings. Table~\ref{tab:casepanel} aligns the finding-level results and report excerpts; Appendix~\ref{app:cases} discusses anatomical details. The aggregate comparison above establishes the same pattern across the validation population.

\subsection{Coupling finding decisions with positive-answer weighting}\label{sec:components}
Explicit finding decisions and positive-answer weighting improve both precision and recall when trained together (Table~\ref{tab:factorial}). Their combination gains \N{twinGap} report-label macro-F1 (\N{twinGapLow} to \N{twinGapHigh}) over the matched state-first control. The component experiment tests two explanations for this result: whether an exhaustive decision prefix is sufficient, and whether the same costs could work by emphasizing finding mentions in free text.

\begin{table}[t]
\centering\small
\caption{Component controls, continuing from the same state-first checkpoint for 649 updates. Every condition retains descriptive states and reports. Costs weight positive finding names without a prefix and positive answers with a prefix. F1, precision (P), and recall (R) are macro-averaged report-label scores on the corrected population.}
\label{tab:factorial}
\begin{tabular}{@{}ccrrr@{}}
\toprule
Decision prefix & Finding-specific costs & F1 & P & R \\
\midrule
-- & -- & \N{twinF} & \N{opTwinP} & \N{opTwinR} \\
-- & $\checkmark$ & \N{rwF} & \N{rwP} & \N{rwR} \\
$\checkmark$ & -- & \N{uwF} & \N{uwP} & \N{uwR} \\
$\checkmark$ & $\checkmark$ & \textbf{\N{pilotF}} & \N{opPilotP} & \N{opPilotR} \\
\bottomrule
\end{tabular}
\end{table}

The two isolated components move along different precision--recall trade-offs. Unweighted decisions increase precision but suppress recall. Weighting finding names in free-text states raises recall while reducing precision, with a macro-F1 change of \N{rwVsTwin} (\N{rwVsTwinLow} to \N{rwVsTwinHigh}) relative to the control. Thus neither an explicit checklist alone nor greater emphasis on finding-related words accounts for the combined improvement. The answer token supplies a well-defined decision location at which the positive cost can act.

Answer probabilities explain why this location matters. The unweighted and weighted decision prefixes have similar ranking performance: their macro-AUROC difference is \N{ssQGap} (unweighted minus weighted; \N{ssQGapLow} to \N{ssQGapHigh}). At the default yes/no boundary, their decision macro-F1 against the reference labels is \N{ssHalfUw} and \N{ssHalfW}; cross-fitted thresholds recover \N{ssCrossUw} and \N{ssCrossW}, respectively. Much of the information absent from the unweighted model's answers at the default boundary remains in its probabilities. For the weighted prefix, the default boundary already achieves nearly the same decision F1 as cross-fitted thresholds. Weighting makes that information available as affirmative generated decisions, which the report can then use. Hiatal hernia also benefits in ranking---removing the costs changes its answer AUROC by \N{ssHiatal}---so the effect includes a finding-specific discrimination gain. Unlike thresholding probabilities after generation, training the answer boundary changes the context on which the report is written.

\subsection{Decisions link visual evidence to report content}\label{sec:intervention}
Agreement between a decision prefix and its report could arise because both respond to the same CT. To test whether the prefix itself governs the report, we edit one answer in the model's generated prefix, keep the CT fixed, and resume decoding. A no-to-yes edit yields a positive report-label prediction in \N{flipYes} of cases; after a yes-to-no edit, only \N{flipNo} remain label-positive. In unmodified outputs, \N{follow} of positive decisions have corresponding positive report-label predictions. The report follows the supplied decision even when a deliberately imposed positive contradicts the reference label. The prefix is therefore an operative interface for report content, rather than merely a correlated intermediate description.

We also test the dependence of report finding recovery on the CT by masking one of six consecutive portions of the processed axial sequence and regenerating the output. For each finding, we measure how often a reference-positive volume with a label-positive baseline report becomes label-negative after masking, and compare this with a second generation from the unchanged CT. Masking the second portion from the caudal end produces \N{occHiatalLoss} report-label loss for hiatal hernia, compared with \N{occHiatalRepeat} under unchanged input. The aggregate responses show finding-dependent spatial losses in report labels. In the paired example, the mask changes the hernia answer from yes to no and removes its report statement; all other masks preserve it. This example traces the input change through both generated outputs. Full masking responses and paired excerpts are given in Appendices~\ref{app:occlusion} and~\ref{app:cases}.

\section{Finding Recovery and Reference-Template Agreement}\label{sec:metrics}
The reversal between finding recovery and text-agreement scores in Table~\ref{tab:report-metrics} raises a concrete question: which aspect of a report changes these metrics? CT generators are known to favor recurring report templates \citep{mayelasserre2026clargen}. We use \emph{template} to denote a recurring report opening and its pattern of normal statements. Matching this pattern can increase agreement with a reference even when the recovered abnormalities are unchanged. We test this explanation with separate interventions on finding presence and template identity.

\paragraph{Supplying finding presence.}
For the shared state-first checkpoint used in the component experiment, we edit its generated descriptive state to match the reference finding labels, then resume report generation from the edited state. We remove rows for reference-negative findings, retain existing reference-positive rows, and add a stock row for each missing positive finding. Other rows are retained. The CT and model weights remain fixed. This intervention supplies the finding content the model should express while leaving its report wording to the decoder.

\paragraph{Supplying reference-template identity.}
We train a full-scale variant to generate a template identifier after its decision prefix and before its descriptive state. At evaluation, we retain each CT's generated decisions and supply the reference-template identifier for the \N{sNamedCTs} volumes whose references use one of nine templates with no benchmark finding in the opening sentence. For the remaining volumes, the model generates its own identifier. The same model also generates freely for the paired comparison. Thus the intervention changes the template information available to the decoder while holding the CT and all prefix answers fixed.

\begin{table}[!ht]
\centering\small
\caption{Paired interventions on finding content and report template, evaluated on all \N{historicalVolumes} volumes. Each cell shows before $\rightarrow$ after. Content is supplied to the shared state-first checkpoint of the component experiment; template identity is supplied to the full-scale template-conditioned variant with its generated decision prefix fixed. F1 is report-label macro-F1; RG is RadGraph-XL complete F1.}
\label{tab:metric-intervention}
\setlength{\tabcolsep}{4pt}
\begin{tabular}{@{}lccccc@{}}
\toprule
Reference input & F1 & RG & BLEU-4 & ROUGE-L & METEOR \\
\midrule
Finding presence & \N{stateF} $\rightarrow$ \N{oracleF} & \N{stateRG} $\rightarrow$ \N{oracleRG} & \N{contentBleuBefore} $\rightarrow$ \N{contentBleuAfter} & \N{contentRougeBefore} $\rightarrow$ \N{contentRougeAfter} & \N{contentMeteorBefore} $\rightarrow$ \N{contentMeteorAfter} \\
Template identity & \N{sFreeF} $\rightarrow$ \N{sNamedFval} & \N{sAllFreeRG} $\rightarrow$ \N{sAllRG} & \N{templateBleuBefore} $\rightarrow$ \N{templateBleuAfter} & \N{templateRougeBefore} $\rightarrow$ \N{templateRougeAfter} & \N{templateMeteorBefore} $\rightarrow$ \N{templateMeteorAfter} \\
\bottomrule
\end{tabular}
\end{table}

The interventions produce complementary responses (Table~\ref{tab:metric-intervention}). Supplying finding presence nearly saturates report-label F1 but changes text-agreement scores modestly. Supplying template identity raises every text-agreement score substantially while report-label F1 changes by only \N{sNamedFPrecise}. On the volumes receiving an identifier, the RadGraph-XL gain is \N{sNamedGain} (\N{sNamedGainLow} to \N{sNamedGainHigh}). The effect extends to descriptions of abnormalities: recall of reference graph items in affirmative finding sentences increases by \N{sAffirmedItems}. Reference-template agreement therefore affects how abnormalities are expressed as well as the normal background text.

\paragraph{Template sensitivity in an LLM judge.}
GREEN compares a candidate with its reference by listing matched findings and errors \citep{ostmeier2024green}. We test its response in a separate intervention on the decision-first model: we hold the generated decision prefix and descriptive state fixed, supply a training-derived opening for the reference's template on the same eligible volumes, and resume generation. This opening is the most frequent first sentence of that template in the training set; the other volumes continue without a supplied opening. Across all validation volumes, GREEN increases by \N{greenOpeningGain} (\N{greenOpeningLow} to \N{greenOpeningHigh}) despite the unchanged decisions. Of the items in the judge's matched-finding lists for the decision-first model, \N{greenMainNormalShare} are statements of normality. A template can consequently improve the judge's score by increasing agreement on normal statements, even when abnormality recovery changes very little. This complements prior rephrasing analyses \citep{sharma2026ctestmetric}: changing a template can change which normal statements appear, as well as their wording.

These results explain why better finding recovery need not produce a higher text-agreement score. Both content and expression matter for reports, but their contributions to a metric must be distinguished. Reporting finding recovery alongside text agreement, with controlled template comparisons where relevant, makes improvements in CT interpretation easier to identify.

\section{Related Work}\label{sec:related}
\paragraph{CT reporting and general video language models.}
CT2Rep established autoregressive volumetric report generation; CT-CHAT and M3D extended CT vision--language learning toward general instruction following \citep{hamamci2024ct2rep,hamamci2026ctrate,bai2024m3d}. AbSteering adapts a general video language model through abnormality-first generation and preference optimization \citep{fang2026absteering}. NV-Reason-CT combines native 3D encoding, additional CT corpora, reasoning supervision, and reinforcement learning \citep{nvreasonct}. Our work develops an autoregressive reporting method on a general video language model, and uses matched generation targets to explain its performance under a common architecture and data budget.

\paragraph{Planning and diagnostic supervision.}
Content planning and observation-guided reporting provide intermediate structure before report generation \citep{hou2023organ,nishino2022coplan}. PromptMRG turns a disease-classification branch into diagnostic prompts, while DCP-PD supplies external discriminative cues at inference \citep{jin2024promptmrg,wang2026dcppd}. Diagnostic Loss and TG-CTRG use asymmetric or finding-focused objectives \citep{dloss,kyung2026tgctrg}. DeCAD studies a visual question answering (VQA)--report gap and generates positive diagnostic assessments before regional report descriptions \citep{wang2026decad}. Our contribution connects intermediate structure and finding-focused supervision: exhaustive finding decisions are generated by the report decoder, and their positive-answer costs shape the context used to produce the report. The matched component experiment distinguishes this formulation from weighting finding mentions in free text.

Recent agentic systems separate perception and report construction further. Rad-VL-JEPA connects discriminative visual question answering, query selection, and report editing; RuleBank learns when independent classifiers and image measurements should trigger report revisions \citep{wang2026decoupling,liang2026rulebank}. Our method learns the transition from finding decisions to report text within one autoregressive decoder.

\paragraph{Recoverable information and generated decisions.}
Hidden-state readouts reveal information that language and vision--language models fail to express in their answers \citep{fu2025hiddenplainsight,orgad2024knowmore}. SelfCue establishes a related discrepancy in CT reporting and addresses it through probe-conditioned decoding followed by distillation \citep{selfcue}. Our readout serves as an analytical instrument: matched generation targets test how the training objective determines which recoverable findings reach the report. The method acts during training on decisions that the report decoder itself generates and conditions on.

\paragraph{Report evaluation.}
The gap between textual overlap and diagnostic content has long been recognized in radiology reporting \citep{boag2020baselines,yu2023radcliq}. RadGraph-XL measures entity and relation agreement, and GREEN uses an LLM to identify matched findings and errors \citep{delbrouck2024radgraphxl,ostmeier2024green}. CT-specific work examines template collapse and metric robustness to rephrasing and factual perturbations \citep{mayelasserre2026clargen,sharma2026ctestmetric}. Our paired interventions separately change supplied finding content and reference-template information, explaining how these factors contribute to scores on the same generated-report task.

\section{Discussion}\label{sec:discussion}
The central engineering result is that a general video language model can become a leading CT report generator by changing how its own decoder learns to select report content. Generating decisions before descriptive text teaches the decoder which findings to carry into its report, while positive-answer weighting shapes the boundary at which it affirms a finding. The component controls and decision edits explain why this design affects the final report: the model learns both which answers to commit to and how to carry those answers into prose. Explicit finding decisions give domain adaptation a place to specify reporting behavior that free-text supervision leaves implicit. The resulting generator needs neither a separate diagnostic model nor an editing stage at inference.

The readout analysis also changes how a weak report should be interpreted. An omitted finding can reflect a failure to use recoverable visual information, and changing the reporting target can address that failure while also improving the visual representation. The relevant design question is therefore how perception and content selection are trained together. The measured gap here concerns CT-RATE's 18 report-derived findings. The paired examples show that recovering those findings does not settle their location or extent, nor the completeness of the report outside the vocabulary. Extending explicit decisions to such attributes is a natural next step, using the same connection between a supervised decision and the description it conditions.

The evaluation experiments further distinguish finding recovery from reference-template agreement. A report may improve in abnormality recovery without moving closer to the reference's reporting conventions; conversely, reference-template information can raise text-agreement scores with little change in finding recovery. Strong reporting systems should be assessed on the findings they recover and the detail with which they describe them, alongside their linguistic quality. Separating these dimensions directs model development toward recovering abnormalities and describing them accurately, while making the effect of reporting conventions visible.

\section{Conclusion}\label{sec:conclusion}
The full-scale decision-first generator closes the measured perception--report gap and establishes state-of-the-art single-model report-label performance on the evaluated CT-RATE protocols using supervised fine-tuning alone. The design principle is to supervise finding selection where it becomes context for the report: explicit decisions and their learning objective jointly determine which abnormalities reach the generated text. Evaluating those recovered findings separately from reference-template agreement makes this improvement visible. Code and model weights will be released to support further development of decision-first CT reporting.

\bibliographystyle{plainnat}
\bibliography{refs}
\clearpage
\input{appendix}
\end{document}

%% file: numbers.tex
\newcommand{\N}[1]{\csname n@#1\endcsname}
\expandafter\def\csname n@calibrationPatients\endcsname{16,511}
\expandafter\def\csname n@calibrationVolumes\endcsname{27,828}
\expandafter\def\csname n@evalPatients\endcsname{1,304}
\expandafter\def\csname n@evalVolumes\endcsname{3,002}
\expandafter\def\csname n@flipAbsentAfter\endcsname{There are millimetric nonspecific nodules in both lungs.}
\expandafter\def\csname n@flipAbsentBefore\endcsname{When examined in the lung parenchyma window; no mass nodule-infiltration was detected in both lung parenchyma.}
\expandafter\def\csname n@flipNo\endcsname{5.4\%}
\expandafter\def\csname n@flipSiblingAfter\endcsname{Calcific atheroma plaques are observed in the coronary arteries.}
\expandafter\def\csname n@flipSiblingBefore\endcsname{Calcific atheroma plaques are observed in the aorta and anterior descending coronary artery.}
\expandafter\def\csname n@flipYes\endcsname{92\%}
\expandafter\def\csname n@follow\endcsname{97\%}
\expandafter\def\csname n@historicalVolumes\endcsname{3,039}
\expandafter\def\csname n@mainCRG\endcsname{0.5596}
\expandafter\def\csname n@benchmarkCRG-direct\endcsname{0.4236}
\expandafter\def\csname n@benchmarkCRG-state\endcsname{0.4966}
\expandafter\def\csname n@mainF\endcsname{0.628}
\expandafter\def\csname n@mainHeadAUROC\endcsname{0.885}
\expandafter\def\csname n@mainHigh\endcsname{0.640}
\expandafter\def\csname n@mainLow\endcsname{0.614}
\expandafter\def\csname n@noScreenF\endcsname{0.545}
\expandafter\def\csname n@noScreenHeadAUROC\endcsname{0.871}
\expandafter\def\csname n@opPilotP\endcsname{0.556}
\expandafter\def\csname n@opPilotR\endcsname{0.664}
\expandafter\def\csname n@opTwinP\endcsname{0.524}
\expandafter\def\csname n@opTwinR\endcsname{0.594}
\expandafter\def\csname n@oracleF\endcsname{0.9594}
\expandafter\def\csname n@oracleRG\endcsname{0.217}
\expandafter\def\csname n@pfMain-00\endcsname{0.678}
\expandafter\def\csname n@pfMain-01\endcsname{0.824}
\expandafter\def\csname n@pfMain-02\endcsname{0.660}
\expandafter\def\csname n@pfMain-03\endcsname{0.555}
\expandafter\def\csname n@pfMain-04\endcsname{0.809}
\expandafter\def\csname n@pfMain-05\endcsname{0.542}
\expandafter\def\csname n@pfMain-06\endcsname{0.557}
\expandafter\def\csname n@pfMain-07\endcsname{0.587}
\expandafter\def\csname n@pfMain-08\endcsname{0.587}
\expandafter\def\csname n@pfMain-09\endcsname{0.718}
\expandafter\def\csname n@pfMain-10\endcsname{0.765}
\expandafter\def\csname n@pfMain-11\endcsname{0.490}
\expandafter\def\csname n@pfMain-12\endcsname{0.847}
\expandafter\def\csname n@pfMain-13\endcsname{0.531}
\expandafter\def\csname n@pfMain-14\endcsname{0.469}
\expandafter\def\csname n@pfMain-15\endcsname{0.705}
\expandafter\def\csname n@pfMain-16\endcsname{0.527}
\expandafter\def\csname n@pfMain-17\endcsname{0.461}
\expandafter\def\csname n@pfNoScreen-00\endcsname{0.668}
\expandafter\def\csname n@pfNoScreen-01\endcsname{0.779}
\expandafter\def\csname n@pfNoScreen-02\endcsname{0.654}
\expandafter\def\csname n@pfNoScreen-03\endcsname{0.369}
\expandafter\def\csname n@pfNoScreen-04\endcsname{0.777}
\expandafter\def\csname n@pfNoScreen-05\endcsname{0.294}
\expandafter\def\csname n@pfNoScreen-06\endcsname{0.393}
\expandafter\def\csname n@pfNoScreen-07\endcsname{0.525}
\expandafter\def\csname n@pfNoScreen-08\endcsname{0.504}
\expandafter\def\csname n@pfNoScreen-09\endcsname{0.651}
\expandafter\def\csname n@pfNoScreen-10\endcsname{0.735}
\expandafter\def\csname n@pfNoScreen-11\endcsname{0.406}
\expandafter\def\csname n@pfNoScreen-12\endcsname{0.818}
\expandafter\def\csname n@pfNoScreen-13\endcsname{0.390}
\expandafter\def\csname n@pfNoScreen-14\endcsname{0.381}
\expandafter\def\csname n@pfNoScreen-15\endcsname{0.640}
\expandafter\def\csname n@pfNoScreen-16\endcsname{0.405}
\expandafter\def\csname n@pfNoScreen-17\endcsname{0.426}
\expandafter\def\csname n@pfP-00\endcsname{0.58}
\expandafter\def\csname n@pfP-01\endcsname{0.79}
\expandafter\def\csname n@pfP-02\endcsname{0.55}
\expandafter\def\csname n@pfP-03\endcsname{0.46}
\expandafter\def\csname n@pfP-04\endcsname{0.76}
\expandafter\def\csname n@pfP-05\endcsname{0.47}
\expandafter\def\csname n@pfP-06\endcsname{0.48}
\expandafter\def\csname n@pfP-07\endcsname{0.60}
\expandafter\def\csname n@pfP-08\endcsname{0.47}
\expandafter\def\csname n@pfP-09\endcsname{0.67}
\expandafter\def\csname n@pfP-10\endcsname{0.76}
\expandafter\def\csname n@pfP-11\endcsname{0.52}
\expandafter\def\csname n@pfP-12\endcsname{0.85}
\expandafter\def\csname n@pfP-13\endcsname{0.46}
\expandafter\def\csname n@pfP-14\endcsname{0.36}
\expandafter\def\csname n@pfP-15\endcsname{0.66}
\expandafter\def\csname n@pfP-16\endcsname{0.45}
\expandafter\def\csname n@pfP-17\endcsname{0.43}
\expandafter\def\csname n@pfR-00\endcsname{0.81}
\expandafter\def\csname n@pfR-01\endcsname{0.86}
\expandafter\def\csname n@pfR-02\endcsname{0.83}
\expandafter\def\csname n@pfR-03\endcsname{0.71}
\expandafter\def\csname n@pfR-04\endcsname{0.87}
\expandafter\def\csname n@pfR-05\endcsname{0.65}
\expandafter\def\csname n@pfR-06\endcsname{0.67}
\expandafter\def\csname n@pfR-07\endcsname{0.58}
\expandafter\def\csname n@pfR-08\endcsname{0.77}
\expandafter\def\csname n@pfR-09\endcsname{0.78}
\expandafter\def\csname n@pfR-10\endcsname{0.77}
\expandafter\def\csname n@pfR-11\endcsname{0.46}
\expandafter\def\csname n@pfR-12\endcsname{0.85}
\expandafter\def\csname n@pfR-13\endcsname{0.62}
\expandafter\def\csname n@pfR-14\endcsname{0.69}
\expandafter\def\csname n@pfR-15\endcsname{0.76}
\expandafter\def\csname n@pfR-16\endcsname{0.63}
\expandafter\def\csname n@pfR-17\endcsname{0.49}
\expandafter\def\csname n@pfWeight-00\endcsname{2.29}
\expandafter\def\csname n@pfWeight-01\endcsname{1.45}
\expandafter\def\csname n@pfWeight-02\endcsname{2.55}
\expandafter\def\csname n@pfWeight-03\endcsname{5.40}
\expandafter\def\csname n@pfWeight-04\endcsname{1.44}
\expandafter\def\csname n@pfWeight-05\endcsname{4.88}
\expandafter\def\csname n@pfWeight-06\endcsname{2.89}
\expandafter\def\csname n@pfWeight-07\endcsname{2.49}
\expandafter\def\csname n@pfWeight-08\endcsname{2.60}
\expandafter\def\csname n@pfWeight-09\endcsname{1.67}
\expandafter\def\csname n@pfWeight-10\endcsname{1.77}
\expandafter\def\csname n@pfWeight-11\endcsname{2.57}
\expandafter\def\csname n@pfWeight-12\endcsname{1.18}
\expandafter\def\csname n@pfWeight-13\endcsname{2.39}
\expandafter\def\csname n@pfWeight-14\endcsname{4.02}
\expandafter\def\csname n@pfWeight-15\endcsname{2.30}
\expandafter\def\csname n@pfWeight-16\endcsname{3.84}
\expandafter\def\csname n@pfWeight-17\endcsname{3.23}
\expandafter\def\csname n@pilotF\endcsname{0.592}
\expandafter\def\csname n@rateCross-main\endcsname{0.628}
\expandafter\def\csname n@rateCross-noScreen\endcsname{0.614}
\expandafter\def\csname n@rateCross-reportOnly\endcsname{0.603}
\expandafter\def\csname n@rateHead-main\endcsname{0.627}
\expandafter\def\csname n@rateHead-noScreen\endcsname{0.601}
\expandafter\def\csname n@rateHead-reportOnly\endcsname{0.476}
\expandafter\def\csname n@rateLoss-main\endcsname{$-$0.001}
\expandafter\def\csname n@rateLoss-noScreen\endcsname{+0.055}
\expandafter\def\csname n@rateLossHigh-main\endcsname{+0.006}
\expandafter\def\csname n@rateLossHigh-noScreen\endcsname{+0.067}
\expandafter\def\csname n@rateLossLow-main\endcsname{$-$0.008}
\expandafter\def\csname n@rateLossLow-noScreen\endcsname{+0.043}
\expandafter\def\csname n@reportOnlyF\endcsname{0.393}
\expandafter\def\csname n@reportOnlyHeadAUROC\endcsname{0.870}
\expandafter\def\csname n@rwF\endcsname{0.539}
\expandafter\def\csname n@rwVsTwin\endcsname{$-$0.003}
\expandafter\def\csname n@rwVsTwinHigh\endcsname{+0.008}
\expandafter\def\csname n@rwVsTwinLow\endcsname{$-$0.013}
\expandafter\def\csname n@sAffirmedItems\endcsname{+0.039}
\expandafter\def\csname n@sAllGain\endcsname{+0.109}
\expandafter\def\csname n@sFreeRG\endcsname{0.248}
\expandafter\def\csname n@sNamedCTs\endcsname{1,922}
\expandafter\def\csname n@sNamedFPrecise\endcsname{+0.0002}
\expandafter\def\csname n@sNamedGain\endcsname{+0.172}
\expandafter\def\csname n@sNamedGainHigh\endcsname{+0.185}
\expandafter\def\csname n@sNamedGainLow\endcsname{+0.159}
\expandafter\def\csname n@sNamedRG\endcsname{0.420}
\expandafter\def\csname n@selectedUpdate\endcsname{4,278}
\expandafter\def\csname n@sideBoth\endcsname{51\%}
\expandafter\def\csname n@sideCases\endcsname{901}
\expandafter\def\csname n@ssCrossUw\endcsname{0.610}
\expandafter\def\csname n@ssCrossW\endcsname{0.614}
\expandafter\def\csname n@ssHiatal\endcsname{$-$0.038}
\expandafter\def\csname n@ssQGap\endcsname{$-$0.0016}
\expandafter\def\csname n@ssQGapHigh\endcsname{+0.0002}
\expandafter\def\csname n@ssQGapLow\endcsname{$-$0.0032}
\expandafter\def\csname n@stateF\endcsname{0.5395}
\expandafter\def\csname n@stateRG\endcsname{0.194}
\expandafter\def\csname n@tmRG-main\endcsname{0.191}
\expandafter\def\csname n@tmRG-noScreen\endcsname{0.193}
\expandafter\def\csname n@tmRG-reportOnly\endcsname{0.237}
\expandafter\def\csname n@trainPatients\endcsname{19,999}
\expandafter\def\csname n@trainStudies\endcsname{24,113}
\expandafter\def\csname n@trainVolumes\endcsname{47,118}
\expandafter\def\csname n@twinF\endcsname{0.542}
\expandafter\def\csname n@twinGap\endcsname{+0.051}
\expandafter\def\csname n@twinGapHigh\endcsname{+0.061}
\expandafter\def\csname n@twinGapLow\endcsname{+0.041}
\expandafter\def\csname n@uwF\endcsname{0.505}
\expandafter\def\csname n@rwP\endcsname{0.504}
\expandafter\def\csname n@rwR\endcsname{0.604}
\expandafter\def\csname n@uwP\endcsname{0.680}
\expandafter\def\csname n@uwR\endcsname{0.435}
\expandafter\def\csname n@sFreeF\endcsname{0.6239}
\expandafter\def\csname n@sNamedFval\endcsname{0.6241}
\expandafter\def\csname n@sAllFreeRG\endcsname{0.213}
\expandafter\def\csname n@sAllRG\endcsname{0.322}
\expandafter\def\csname n@occHiatalLoss\endcsname{44.2\%}
\expandafter\def\csname n@occCardiacLoss\endcsname{39.5\%}
\expandafter\def\csname n@occHiatalRepeat\endcsname{0.0\%}
\expandafter\def\csname n@occCardiacRepeat\endcsname{0.4\%}
\expandafter\def\csname n@caseEligible\endcsname{11}
\expandafter\def\csname n@casePanelVolume\endcsname{\detokenize{valid_806_a_1}}
\expandafter\def\csname n@casePanelReferenceCount\endcsname{4}
\expandafter\def\csname n@casePanelReferenceA\endcsname{Medical material}
\expandafter\def\csname n@casePanelReferenceB\endcsname{Arterial wall calcification}
\expandafter\def\csname n@casePanelReferenceC\endcsname{Coronary artery wall calcification}
\expandafter\def\csname n@casePanelReferenceD\endcsname{Lung nodule}
\expandafter\def\csname n@casePanelDirectExtraA\endcsname{Hiatal hernia}
\expandafter\def\csname n@casePanelDirectExtraB\endcsname{Emphysema}
\expandafter\def\csname n@occHeatN00\endcsname{250}
\expandafter\def\csname n@occHeat00P1\endcsname{4.4}
\expandafter\def\csname n@occHeat00P2\endcsname{2.4}
\expandafter\def\csname n@occHeat00P3\endcsname{5.6}
\expandafter\def\csname n@occHeat00P4\endcsname{11.6}
\expandafter\def\csname n@occHeat00P5\endcsname{7.2}
\expandafter\def\csname n@occHeat00P6\endcsname{2.8}
\expandafter\def\csname n@occHeatR00\endcsname{0.4}
\expandafter\def\csname n@occHeatN01\endcsname{734}
\expandafter\def\csname n@occHeat01P1\endcsname{1.6}
\expandafter\def\csname n@occHeat01P2\endcsname{1.1}
\expandafter\def\csname n@occHeat01P3\endcsname{4.0}
\expandafter\def\csname n@occHeat01P4\endcsname{11.7}
\expandafter\def\csname n@occHeat01P5\endcsname{15.5}
\expandafter\def\csname n@occHeat01P6\endcsname{0.5}
\expandafter\def\csname n@occHeatR01\endcsname{0.4}
\expandafter\def\csname n@occHeatN02\endcsname{261}
\expandafter\def\csname n@occHeat02P1\endcsname{1.9}
\expandafter\def\csname n@occHeat02P2\endcsname{1.5}
\expandafter\def\csname n@occHeat02P3\endcsname{39.5}
\expandafter\def\csname n@occHeat02P4\endcsname{5.4}
\expandafter\def\csname n@occHeat02P5\endcsname{1.1}
\expandafter\def\csname n@occHeat02P6\endcsname{0.0}
\expandafter\def\csname n@occHeatR02\endcsname{0.4}
\expandafter\def\csname n@occHeatN03\endcsname{154}
\expandafter\def\csname n@occHeat03P1\endcsname{1.3}
\expandafter\def\csname n@occHeat03P2\endcsname{10.4}
\expandafter\def\csname n@occHeat03P3\endcsname{24.7}
\expandafter\def\csname n@occHeat03P4\endcsname{7.8}
\expandafter\def\csname n@occHeat03P5\endcsname{0.6}
\expandafter\def\csname n@occHeat03P6\endcsname{2.6}
\expandafter\def\csname n@occHeatR03\endcsname{1.3}
\expandafter\def\csname n@occHeatN04\endcsname{660}
\expandafter\def\csname n@occHeat04P1\endcsname{1.4}
\expandafter\def\csname n@occHeat04P2\endcsname{0.9}
\expandafter\def\csname n@occHeat04P3\endcsname{16.2}
\expandafter\def\csname n@occHeat04P4\endcsname{30.5}
\expandafter\def\csname n@occHeat04P5\endcsname{6.2}
\expandafter\def\csname n@occHeat04P6\endcsname{0.8}
\expandafter\def\csname n@occHeatR04\endcsname{0.2}
\expandafter\def\csname n@occHeatN05\endcsname{269}
\expandafter\def\csname n@occHeat05P1\endcsname{5.2}
\expandafter\def\csname n@occHeat05P2\endcsname{44.2}
\expandafter\def\csname n@occHeat05P3\endcsname{21.6}
\expandafter\def\csname n@occHeat05P4\endcsname{3.7}
\expandafter\def\csname n@occHeat05P5\endcsname{2.2}
\expandafter\def\csname n@occHeat05P6\endcsname{1.9}
\expandafter\def\csname n@occHeatR05\endcsname{0.0}
\expandafter\def\csname n@occHeatN06\endcsname{519}
\expandafter\def\csname n@occHeat06P1\endcsname{1.7}
\expandafter\def\csname n@occHeat06P2\endcsname{0.0}
\expandafter\def\csname n@occHeat06P3\endcsname{8.3}
\expandafter\def\csname n@occHeat06P4\endcsname{21.8}
\expandafter\def\csname n@occHeat06P5\endcsname{6.9}
\expandafter\def\csname n@occHeat06P6\endcsname{1.0}
\expandafter\def\csname n@occHeatR06\endcsname{1.3}
\expandafter\def\csname n@occHeatN07\endcsname{340}
\expandafter\def\csname n@occHeat07P1\endcsname{0.9}
\expandafter\def\csname n@occHeat07P2\endcsname{4.7}
\expandafter\def\csname n@occHeat07P3\endcsname{4.4}
\expandafter\def\csname n@occHeat07P4\endcsname{7.9}
\expandafter\def\csname n@occHeat07P5\endcsname{14.7}
\expandafter\def\csname n@occHeat07P6\endcsname{10.6}
\expandafter\def\csname n@occHeatR07\endcsname{0.6}
\expandafter\def\csname n@occHeatN08\endcsname{542}
\expandafter\def\csname n@occHeat08P1\endcsname{3.5}
\expandafter\def\csname n@occHeat08P2\endcsname{10.0}
\expandafter\def\csname n@occHeat08P3\endcsname{21.8}
\expandafter\def\csname n@occHeat08P4\endcsname{5.4}
\expandafter\def\csname n@occHeat08P5\endcsname{4.1}
\expandafter\def\csname n@occHeat08P6\endcsname{2.4}
\expandafter\def\csname n@occHeatR08\endcsname{1.8}
\expandafter\def\csname n@occHeatN09\endcsname{1,055}
\expandafter\def\csname n@occHeat09P1\endcsname{2.3}
\expandafter\def\csname n@occHeat09P2\endcsname{5.5}
\expandafter\def\csname n@occHeat09P3\endcsname{12.0}
\expandafter\def\csname n@occHeat09P4\endcsname{18.4}
\expandafter\def\csname n@occHeat09P5\endcsname{9.8}
\expandafter\def\csname n@occHeat09P6\endcsname{2.5}
\expandafter\def\csname n@occHeatR09\endcsname{1.1}
\expandafter\def\csname n@occHeatN10\endcsname{901}
\expandafter\def\csname n@occHeat10P1\endcsname{4.0}
\expandafter\def\csname n@occHeat10P2\endcsname{5.1}
\expandafter\def\csname n@occHeat10P3\endcsname{10.2}
\expandafter\def\csname n@occHeat10P4\endcsname{7.8}
\expandafter\def\csname n@occHeat10P5\endcsname{6.4}
\expandafter\def\csname n@occHeat10P6\endcsname{4.3}
\expandafter\def\csname n@occHeatR10\endcsname{2.1}
\expandafter\def\csname n@occHeatN11\endcsname{381}
\expandafter\def\csname n@occHeat11P1\endcsname{7.1}
\expandafter\def\csname n@occHeat11P2\endcsname{9.2}
\expandafter\def\csname n@occHeat11P3\endcsname{16.0}
\expandafter\def\csname n@occHeat11P4\endcsname{18.4}
\expandafter\def\csname n@occHeat11P5\endcsname{24.9}
\expandafter\def\csname n@occHeat11P6\endcsname{34.4}
\expandafter\def\csname n@occHeatR11\endcsname{3.9}
\expandafter\def\csname n@occHeatN12\endcsname{313}
\expandafter\def\csname n@occHeat12P1\endcsname{0.3}
\expandafter\def\csname n@occHeat12P2\endcsname{4.8}
\expandafter\def\csname n@occHeat12P3\endcsname{7.0}
\expandafter\def\csname n@occHeat12P4\endcsname{2.2}
\expandafter\def\csname n@occHeat12P5\endcsname{0.6}
\expandafter\def\csname n@occHeat12P6\endcsname{0.3}
\expandafter\def\csname n@occHeatR12\endcsname{0.6}
\expandafter\def\csname n@occHeatN13\endcsname{152}
\expandafter\def\csname n@occHeat13P1\endcsname{3.3}
\expandafter\def\csname n@occHeat13P2\endcsname{5.3}
\expandafter\def\csname n@occHeat13P3\endcsname{19.1}
\expandafter\def\csname n@occHeat13P4\endcsname{14.5}
\expandafter\def\csname n@occHeat13P5\endcsname{7.2}
\expandafter\def\csname n@occHeat13P6\endcsname{1.3}
\expandafter\def\csname n@occHeatR13\endcsname{1.3}
\expandafter\def\csname n@occHeatN14\endcsname{240}
\expandafter\def\csname n@occHeat14P1\endcsname{1.7}
\expandafter\def\csname n@occHeat14P2\endcsname{4.6}
\expandafter\def\csname n@occHeat14P3\endcsname{27.9}
\expandafter\def\csname n@occHeat14P4\endcsname{15.8}
\expandafter\def\csname n@occHeat14P5\endcsname{7.9}
\expandafter\def\csname n@occHeat14P6\endcsname{5.0}
\expandafter\def\csname n@occHeatR14\endcsname{1.7}
\expandafter\def\csname n@occHeatN15\endcsname{440}
\expandafter\def\csname n@occHeat15P1\endcsname{3.4}
\expandafter\def\csname n@occHeat15P2\endcsname{5.9}
\expandafter\def\csname n@occHeat15P3\endcsname{10.7}
\expandafter\def\csname n@occHeat15P4\endcsname{13.0}
\expandafter\def\csname n@occHeat15P5\endcsname{8.0}
\expandafter\def\csname n@occHeat15P6\endcsname{2.3}
\expandafter\def\csname n@occHeatR15\endcsname{1.6}
\expandafter\def\csname n@occHeatN16\endcsname{206}
\expandafter\def\csname n@occHeat16P1\endcsname{2.4}
\expandafter\def\csname n@occHeat16P2\endcsname{5.8}
\expandafter\def\csname n@occHeat16P3\endcsname{25.7}
\expandafter\def\csname n@occHeat16P4\endcsname{15.5}
\expandafter\def\csname n@occHeat16P5\endcsname{14.1}
\expandafter\def\csname n@occHeat16P6\endcsname{2.4}
\expandafter\def\csname n@occHeatR16\endcsname{2.4}
\expandafter\def\csname n@occHeatN17\endcsname{121}
\expandafter\def\csname n@occHeat17P1\endcsname{1.7}
\expandafter\def\csname n@occHeat17P2\endcsname{15.7}
\expandafter\def\csname n@occHeat17P3\endcsname{17.4}
\expandafter\def\csname n@occHeat17P4\endcsname{14.9}
\expandafter\def\csname n@occHeat17P5\endcsname{11.6}
\expandafter\def\csname n@occHeat17P6\endcsname{2.5}
\expandafter\def\csname n@occHeatR17\endcsname{2.5}
\expandafter\def\csname n@mainFullF\endcsname{0.6250}
\expandafter\def\csname n@mainFPrecise\endcsname{0.6284}
\expandafter\def\csname n@mainFullLow\endcsname{0.611}
\expandafter\def\csname n@mainFullHigh\endcsname{0.638}
\expandafter\def\csname n@fullGain\endcsname{+0.083}
\expandafter\def\csname n@fullGainLow\endcsname{+0.071}
\expandafter\def\csname n@fullGainHigh\endcsname{+0.095}
\expandafter\def\csname n@benchmarkCorrected-direct\endcsname{0.3929}
\expandafter\def\csname n@benchmarkCorrected-state\endcsname{0.5453}
\expandafter\def\csname n@qualityFullF-direct\endcsname{0.3907}
\expandafter\def\csname n@qualityBleu-direct\endcsname{0.279}
\expandafter\def\csname n@qualityRouge-direct\endcsname{0.412}
\expandafter\def\csname n@qualityMeteor-direct\endcsname{0.438}
\expandafter\def\csname n@readP-direct\endcsname{0.575}
\expandafter\def\csname n@readR-direct\endcsname{0.324}
\expandafter\def\csname n@qualityFullF-state\endcsname{0.5400}
\expandafter\def\csname n@qualityBleu-state\endcsname{0.235}
\expandafter\def\csname n@qualityRouge-state\endcsname{0.364}
\expandafter\def\csname n@qualityMeteor-state\endcsname{0.427}
\expandafter\def\csname n@readP-state\endcsname{0.536}
\expandafter\def\csname n@readR-state\endcsname{0.571}
\expandafter\def\csname n@qualityBleu-main\endcsname{0.237}
\expandafter\def\csname n@qualityRouge-main\endcsname{0.364}
\expandafter\def\csname n@qualityMeteor-main\endcsname{0.435}
\expandafter\def\csname n@readP-main\endcsname{0.573}
\expandafter\def\csname n@readR-main\endcsname{0.710}
\expandafter\def\csname n@contentBleuBefore\endcsname{0.233}
\expandafter\def\csname n@contentBleuAfter\endcsname{0.256}
\expandafter\def\csname n@templateBleuBefore\endcsname{0.263}
\expandafter\def\csname n@templateBleuAfter\endcsname{0.392}
\expandafter\def\csname n@contentRougeBefore\endcsname{0.357}
\expandafter\def\csname n@contentRougeAfter\endcsname{0.381}
\expandafter\def\csname n@templateRougeBefore\endcsname{0.383}
\expandafter\def\csname n@templateRougeAfter\endcsname{0.500}
\expandafter\def\csname n@contentMeteorBefore\endcsname{0.421}
\expandafter\def\csname n@contentMeteorAfter\endcsname{0.444}
\expandafter\def\csname n@templateMeteorBefore\endcsname{0.451}
\expandafter\def\csname n@templateMeteorAfter\endcsname{0.534}
\expandafter\def\csname n@greenOpeningGain\endcsname{+0.047}
\expandafter\def\csname n@greenOpeningLow\endcsname{+0.042}
\expandafter\def\csname n@greenOpeningHigh\endcsname{+0.053}
\expandafter\def\csname n@greenMain\endcsname{0.469}
\expandafter\def\csname n@greenOpening\endcsname{0.517}
\expandafter\def\csname n@greenMainNormalShare\endcsname{88\%}
\expandafter\def\csname n@ssHalfUw\endcsname{0.508}
\expandafter\def\csname n@ssHalfW\endcsname{0.609}
\expandafter\def\csname n@greenOpeningF\endcsname{+0.0012}
\expandafter\def\csname n@occCaseVolume\endcsname{\detokenize{valid_1004_a_1}}
\expandafter\def\csname n@occCaseQuote\endcsname{Sliding type hiatal hernia was observed at the lower end of the esophagus. Impression:
Hiatal hernia.}
\expandafter\def\csname n@flipAbsentVolume\endcsname{\detokenize{valid_1000_a_2}}
\expandafter\def\csname n@flipSiblingVolume\endcsname{\detokenize{valid_101_b_2}}
\expandafter\def\csname n@caseDirectReadoutA\endcsname{1}
\expandafter\def\csname n@caseDirectReadoutB\endcsname{1}
\expandafter\def\csname n@caseDirectReadoutC\endcsname{1}
\expandafter\def\csname n@caseDirectReadoutD\endcsname{1}
\expandafter\def\csname n@caseDirectReportA\endcsname{0}
\expandafter\def\csname n@caseDirectReportB\endcsname{0}
\expandafter\def\csname n@caseDirectReportC\endcsname{0}
\expandafter\def\csname n@caseDirectReportD\endcsname{0}
\expandafter\def\csname n@caseStateReadoutA\endcsname{1}
\expandafter\def\csname n@caseStateReadoutB\endcsname{1}
\expandafter\def\csname n@caseStateReadoutC\endcsname{1}
\expandafter\def\csname n@caseStateReadoutD\endcsname{1}
\expandafter\def\csname n@caseStateReportA\endcsname{1}
\expandafter\def\csname n@caseStateReportB\endcsname{0}
\expandafter\def\csname n@caseStateReportC\endcsname{0}
\expandafter\def\csname n@caseStateReportD\endcsname{0}
\expandafter\def\csname n@caseOursReadoutA\endcsname{1}
\expandafter\def\csname n@caseOursReadoutB\endcsname{1}
\expandafter\def\csname n@caseOursReadoutC\endcsname{1}
\expandafter\def\csname n@caseOursReadoutD\endcsname{1}
\expandafter\def\csname n@caseOursDecisionA\endcsname{1}
\expandafter\def\csname n@caseOursDecisionB\endcsname{1}
\expandafter\def\csname n@caseOursDecisionC\endcsname{1}
\expandafter\def\csname n@caseOursDecisionD\endcsname{1}
\expandafter\def\csname n@caseOursReportA\endcsname{1}
\expandafter\def\csname n@caseOursReportB\endcsname{1}
\expandafter\def\csname n@caseOursReportC\endcsname{1}
\expandafter\def\csname n@caseOursReportD\endcsname{1}
\expandafter\def\csname n@caseReferenceVascular\endcsname{Diffuse calcific atheroma plaques were observed in the arcus-descending aorta and coronary arteries. There is a stent placed in the LAD and Cx.}
\expandafter\def\csname n@caseReferenceNodules\endcsname{A few millimetric nonspecific parenchymal nodules were observed in both lungs.}
\expandafter\def\csname n@caseDirectVascular\endcsname{Mediastinal main vascular structures, heart contour, size are normal.}
\expandafter\def\csname n@caseDirectNodules\endcsname{Aeration of both lung parenchyma is normal and no nodular or infiltrative lesion is detected in the lung parenchyma.}
\expandafter\def\csname n@caseStateVascular\endcsname{Stent materials are observed in the coronary arteries.}
\expandafter\def\csname n@caseStateNodules\endcsname{no mass nodule-infiltration was detected in both lung parenchyma.}
\expandafter\def\csname n@caseOursVascular\endcsname{Calcific atheroma plaques in the aorta and coronary arteries and stent-like appearances in the coronary arteries were observed.}
\expandafter\def\csname n@caseOursNodules\endcsname{Millimetric nonspecific pulmonary nodules were observed in both lungs.}

%% file: figures/method.tex
% CONTENT CONTRACT: author-supplied vector PDF, referenced in paper.json.
% One CT input feeds one decoder; decisions, state, and report are output spans.
% The bottom row contains training targets, connected by dashed supervision arrows.
\begin{figure}[t]
\centering
\includegraphics[width=\linewidth]{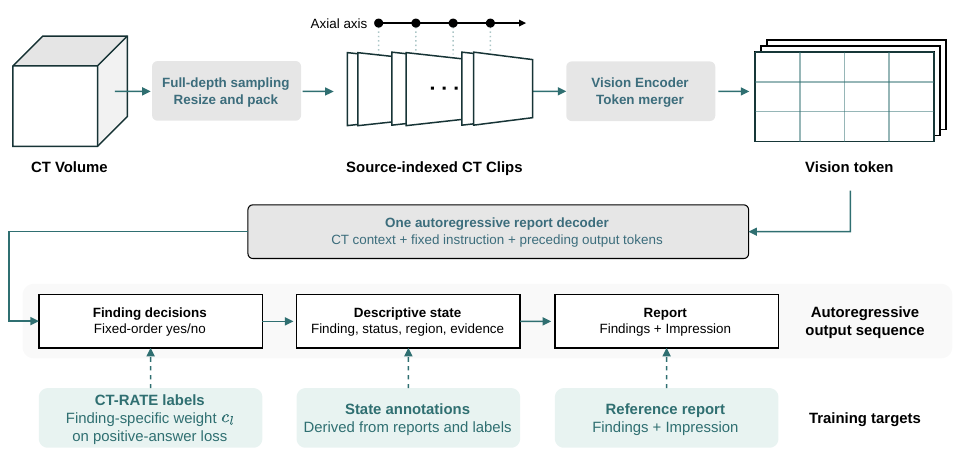}
\caption{CT representation and decision-first generation. Full-depth sampling and clip packing preserve volume coverage, while source-slice indices encode axial order. All clips enter one model input. One decoder generates finding decisions, a descriptive state, and the report as successive parts of the same sequence, retaining CT context throughout. Dashed arrows connect training targets to their output segments; $c_\ell$ weights positive decision tokens during training. At inference, only the CT and fixed instruction are supplied.}
\label{fig:method}
\end{figure}

%% file: appendix.tex
\appendix
\section{Implementation Settings}\label{app:implementation}
Table~\ref{tab:implementation} specifies the shared input and optimization settings. The planned code release will provide preprocessing, target construction, and training configurations.

\begin{table}[h]
\centering\small
\caption{Shared implementation settings. The full-scale models start from public weights; component controls continue from a shared state-first checkpoint.}
\label{tab:implementation}
\begin{tabular}{@{}L{0.28\linewidth}L{0.68\linewidth}@{}}
\toprule
Input & Broad-window grayscale, approximately $[-1024,1024]$ HU; intensities quantized into 256 bins and replicated across three channels. \\
Sampling & Up to 384 uniformly spaced axial slices; $384\times384$ bilinear resize; 32-frame clips with final-slice padding. No augmentation or processor resampling. \\
Positions & Mean source-slice indices for adjacent-frame groups; continuous across clips. No physical-spacing input. \\
Objective & State weight 4, Findings 1, Impression 2; yes-answer weight $4c_\ell$. Findings evidence-span weights: 2 for present/uncertain, 1.5 for absent. Loss is normalized by weighted target-token mass. \\
Optimization & All parameters fine-tuned; bfloat16, AdamW8bit, $\beta=(0.9,0.95)$, weight decay 0.01, batch size 24. Full-scale learning rate $10^{-5}$, multiplied by 0.3 after update 2,620; 100 warm-up updates. \\
Generation & Greedy decoding; maximum 3,072 new tokens. Every generated report is included in evaluation. \\
Visual readout & 18 learned queries, width 512, two cross-attention/feed-forward layers, eight heads; weighted binary cross-entropy; learning rate $10^{-4}$. \\
\bottomrule
\end{tabular}
\end{table}

\paragraph{Cost estimation.}
Four readouts are trained on frozen visual tokens from the shared state-first checkpoint with patient-disjoint folds. Their pooled out-of-fold predictions cover \N{calibrationVolumes} training reconstructions from \N{calibrationPatients} patients; a separate development panel is excluded. For each finding, the F1-maximizing logit threshold $t_\ell$ gives $c_\ell=\exp(-t_\ell)$, rounded to two decimals. This is Equation~\ref{eq:cost} with $\tau_\ell=\sigma(t_\ell)$. Table~\ref{tab:per-finding} lists the fixed costs. The shared state-first checkpoint used for component controls and cost estimation had received self-critical report training with a source-relative lexical reward. The full-scale models start from public weights; only the 18 scalar costs are carried over to decision-first training.

\section{Finding-Level Results}\label{app:findings}
Table~\ref{tab:per-finding} gives finding-specific costs and report scores on the corrected population. On all \N{historicalVolumes} volumes, the decision-first model's report-label macro-F1 has a patient-cluster 95\% interval of \N{mainFullLow}--\N{mainFullHigh}.

\begin{table}[h]
\centering\small
\caption{Finding-specific positive-answer costs and report scores for the matched full-scale models. P and R denote precision and recall.}
\label{tab:per-finding}
\setlength{\tabcolsep}{4pt}
\begin{tabular}{@{}L{0.34\linewidth}rrrrr@{}}
\toprule
Finding & Cost & State-first & \multicolumn{3}{c}{Decision-first} \\
\cmidrule(lr){3-3}\cmidrule(l){4-6}
& & F1 & F1 & P & R \\
\midrule
Medical material & \N{pfWeight-00} & \N{pfNoScreen-00} & \N{pfMain-00} & \N{pfP-00} & \N{pfR-00} \\
Arterial wall calcification & \N{pfWeight-01} & \N{pfNoScreen-01} & \N{pfMain-01} & \N{pfP-01} & \N{pfR-01} \\
Cardiomegaly & \N{pfWeight-02} & \N{pfNoScreen-02} & \N{pfMain-02} & \N{pfP-02} & \N{pfR-02} \\
Pericardial effusion & \N{pfWeight-03} & \N{pfNoScreen-03} & \N{pfMain-03} & \N{pfP-03} & \N{pfR-03} \\
Coronary artery wall calcification & \N{pfWeight-04} & \N{pfNoScreen-04} & \N{pfMain-04} & \N{pfP-04} & \N{pfR-04} \\
Hiatal hernia & \N{pfWeight-05} & \N{pfNoScreen-05} & \N{pfMain-05} & \N{pfP-05} & \N{pfR-05} \\
Lymphadenopathy & \N{pfWeight-06} & \N{pfNoScreen-06} & \N{pfMain-06} & \N{pfP-06} & \N{pfR-06} \\
Emphysema & \N{pfWeight-07} & \N{pfNoScreen-07} & \N{pfMain-07} & \N{pfP-07} & \N{pfR-07} \\
Atelectasis & \N{pfWeight-08} & \N{pfNoScreen-08} & \N{pfMain-08} & \N{pfP-08} & \N{pfR-08} \\
Lung nodule & \N{pfWeight-09} & \N{pfNoScreen-09} & \N{pfMain-09} & \N{pfP-09} & \N{pfR-09} \\
Lung opacity & \N{pfWeight-10} & \N{pfNoScreen-10} & \N{pfMain-10} & \N{pfP-10} & \N{pfR-10} \\
Pulmonary fibrotic sequela & \N{pfWeight-11} & \N{pfNoScreen-11} & \N{pfMain-11} & \N{pfP-11} & \N{pfR-11} \\
Pleural effusion & \N{pfWeight-12} & \N{pfNoScreen-12} & \N{pfMain-12} & \N{pfP-12} & \N{pfR-12} \\
Mosaic attenuation pattern & \N{pfWeight-13} & \N{pfNoScreen-13} & \N{pfMain-13} & \N{pfP-13} & \N{pfR-13} \\
Peribronchial thickening & \N{pfWeight-14} & \N{pfNoScreen-14} & \N{pfMain-14} & \N{pfP-14} & \N{pfR-14} \\
Consolidation & \N{pfWeight-15} & \N{pfNoScreen-15} & \N{pfMain-15} & \N{pfP-15} & \N{pfR-15} \\
Bronchiectasis & \N{pfWeight-16} & \N{pfNoScreen-16} & \N{pfMain-16} & \N{pfP-16} & \N{pfR-16} \\
Interlobular septal thickening & \N{pfWeight-17} & \N{pfNoScreen-17} & \N{pfMain-17} & \N{pfP-17} & \N{pfR-17} \\
\bottomrule
\end{tabular}
\end{table}

\FloatBarrier
\section{Paired Report Examples}\label{app:cases}
\paragraph{Finding recovery within the same CT.}
Table~\ref{tab:casepanel} follows each reference finding from the visual readout to the generated report. The state-first model reports the coronary stent but omits calcification and negates lung nodules; the decision-first model carries all four positive decisions into its report.

\paragraph{Case selection.}
The CT in Table~\ref{tab:casepanel} is selected from corrected validation volumes with at least two positive reference labels, all of them recovered by the direct-report model's cross-fitted visual readout, no positive label extracted from its report, and all reference findings affirmed in our model's decision prefix and exactly recovered in its report. Among \N{caseEligible} eligible volumes, we select the one with the most positive reference labels, breaking ties by lexicographic volume order. The reference also describes hepatic steatosis, outside the decision vocabulary; our report omits it and describes the upper abdominal organs as normal. This case therefore separates successful recovery of the labeled findings from completeness of the full report.

\input{tables/casepanel}

\paragraph{Finding presence and anatomical detail.}
For \texttt{valid\_1017\_a\_1}, the reference describes ``Mild atelectatic changes are observed in the left lung upper lobe inferior lingula'' and ``A few subpleural nonspecific millimetric nodules are observed in both lungs.'' The direct-report control concludes ``Thoracic CT examination within normal limits.'' Our model's prefix affirms atelectasis and lung nodules, and its Findings report ``linear atelectasis in the right lung middle lobe medial segment and left lung lingula inferior segment'' and ``several millimetric nonspecific nodules in both lungs.'' Both labeled abnormalities are recovered, but the description adds a right-middle-lobe location absent from the reference, drops the subpleural qualifier, and leaves atelectasis out of the Impression. This is the first corrected validation volume in lexicographic order with at least two positive reference labels, exact label recovery by our model, and no positive reference label recovered by the direct-report control.

\clearpage
\paragraph{Paired interventions.}
Table~\ref{tab:intervention-cases} aligns the original and post-intervention outputs. Masking changes both the hernia decision and its report statement; imposing a positive nodule answer introduces a finding absent from the reference; changing the arterial-calcification answer to no removes the aortic-calcification statement while preserving coronary calcification. The nodule case is the first true-negative example in lexicographic order whose post-intervention report is label-positive and affirms the imposed finding with a mild qualifier. The arterial example is selected from edits to a sentence mentioning both arterial and coronary calcification where the coronary finding remains present. The masking case is selected as described in Appendix~\ref{app:occlusion}.

\input{tables/intervention-cases}

\paragraph{Finding presence and spatial description.}
Among labeled findings stated by the generated report and localized to one side in the reference, \N{sideBoth} of \N{sideCases} finding--volume cases are described as bilateral. The atelectasis example above illustrates the distinction between recovering a finding and describing its location accurately.

\FloatBarrier
\section{Additional Evaluation Results}\label{app:metric-details}
\paragraph{Template-conditioned subset.}
Among the \N{sNamedCTs} volumes receiving a supported identifier, RadGraph-XL rises from \N{sFreeRG} to \N{sNamedRG}. Over all validation volumes, its change is \N{sAllGain}. Holding the decision prefix fixed allows the report wording and individual extracted labels to change; nearly unchanged aggregate report-label F1 is the observed outcome.

\paragraph{GREEN and finding recovery.}
For the fixed-prefix opening intervention described in Section~\ref{sec:metrics}, GREEN rises from \N{greenMain} to \N{greenOpening} on the original validation population. On the corrected population, report-label macro-F1 changes by \N{greenOpeningF}.

\clearpage
\section{Visual Evidence Under Input Masking}\label{app:occlusion}
The masking experiment uses the full-scale decision-first model at update \N{selectedUpdate} on the corrected \N{evalVolumes} reconstructions. After the shared preprocessing, we concatenate the 8-bit frame clips, including final-frame padding, and divide the displayed axial sequence into six consecutive portions using rounded boundaries. Each intervention sets one portion to zero (black), retaining the sequence length and positional indices. Portions run from caudal to cranial along this displayed sequence. They are portions of the model input, not equal physical lengths of the original anatomy.

For each finding, the denominator comprises reference-positive volumes whose baseline reports are label-positive. Figure~\ref{fig:occlusion} shows the fraction becoming label-negative under each mask. A second generation from the unchanged CT measures repeat-reading variation. The pattern differs by finding: the third-portion mask produces \N{occCardiacLoss} report-label loss for cardiomegaly, compared with \N{occCardiacRepeat} under unchanged repeat reading; hernia loss concentrates in lower portions.

\input{figures/occlusion}

The paired hernia case in Table~\ref{tab:intervention-cases} is the first volume in lexicographic order among reference-positive hernias stated in both unchanged readings and lost only when the second portion is masked. All other masks preserve the finding. This example traces an input change through the generated decision and into the report.

%% file: tables/casepanel.tex
% CONTENT CONTRACT (SPINE C8): same CT and four reference-positive findings.
% A: three models, with separate analytical readout and report-label columns;
% decision-first additionally shows its self-generated yes/no answers.
% Every mark is list membership bound to casepanel.json, never inferred from prose.
% B: exact Findings excerpts aligned by vascular/material content and lung nodules.
% Readout extras are named below A; readouts never condition generation.
\begin{table}[!htbp]
\centering\small
\caption{From recoverable findings to report content on the same CT (\texttt{\N{casePanelVolume}}). The direct-report readout recovers all \N{casePanelReferenceCount} reference findings, while its report expresses none; decision-first generation affirms and reports each one. Panel A identifies the findings at each output, and Panel B shows the corresponding report text.}
\label{tab:casepanel}
\begingroup
\setlength{\tabcolsep}{2.5pt}
\renewcommand{\arraystretch}{1.2}
\newcommand{\caseMark}[1]{\ifnum#1=1\relax $+$\else $-$\fi}
\newcommand{\caseAnswer}[1]{\ifnum#1=1\relax yes\else no\fi}
\noindent\textbf{A. Finding-level recovery}\par\smallskip
\begin{tabular}{@{}L{0.29\linewidth}*{7}{>{\centering\arraybackslash}p{0.083\linewidth}}@{}}
\toprule
& \multicolumn{2}{c}{Direct-report} & \multicolumn{2}{c}{State-first} & \multicolumn{3}{c}{Decision-first} \\
\cmidrule(lr){2-3}\cmidrule(lr){4-5}\cmidrule(l){6-8}
Reference-positive finding & Readout & Report & Readout & Report & Readout & Decision & Report \\
\midrule
\N{casePanelReferenceA} & \caseMark{\N{caseDirectReadoutA}} & \caseMark{\N{caseDirectReportA}} & \caseMark{\N{caseStateReadoutA}} & \caseMark{\N{caseStateReportA}} & \caseMark{\N{caseOursReadoutA}} & \caseAnswer{\N{caseOursDecisionA}} & \caseMark{\N{caseOursReportA}} \\
\N{casePanelReferenceB} & \caseMark{\N{caseDirectReadoutB}} & \caseMark{\N{caseDirectReportB}} & \caseMark{\N{caseStateReadoutB}} & \caseMark{\N{caseStateReportB}} & \caseMark{\N{caseOursReadoutB}} & \caseAnswer{\N{caseOursDecisionB}} & \caseMark{\N{caseOursReportB}} \\
\N{casePanelReferenceC} & \caseMark{\N{caseDirectReadoutC}} & \caseMark{\N{caseDirectReportC}} & \caseMark{\N{caseStateReadoutC}} & \caseMark{\N{caseStateReportC}} & \caseMark{\N{caseOursReadoutC}} & \caseAnswer{\N{caseOursDecisionC}} & \caseMark{\N{caseOursReportC}} \\
\N{casePanelReferenceD} & \caseMark{\N{caseDirectReadoutD}} & \caseMark{\N{caseDirectReportD}} & \caseMark{\N{caseStateReadoutD}} & \caseMark{\N{caseStateReportD}} & \caseMark{\N{caseOursReadoutD}} & \caseAnswer{\N{caseOursDecisionD}} & \caseMark{\N{caseOursReportD}} \\
\bottomrule
\end{tabular}
\par\vspace{4pt}
\begin{minipage}{\linewidth}\footnotesize
$+/-$: positive/negative under the analytical readout or report-label classifier; yes/no: generated decision tokens. Readouts use patient-cross-fitted thresholds and do not condition generation. Reports are scored from Findings and Impression. The direct-report readout additionally calls \N{casePanelDirectExtraA} and \N{casePanelDirectExtraB}, absent from the reference labels.
\end{minipage}

\par\vspace{10pt}
\noindent\textbf{B. Corresponding Findings excerpts}\par\smallskip
\begin{tabular}{@{}L{0.15\linewidth}L{0.445\linewidth}L{0.365\linewidth}@{}}
\toprule
Source & Vascular findings and medical material & Lung nodules \\
\midrule
Reference & ``\N{caseReferenceVascular}'' & ``\N{caseReferenceNodules}'' \\[5pt]
Direct-report & ``\N{caseDirectVascular}'' & ``\N{caseDirectNodules}'' \\[5pt]
State-first & ``\N{caseStateVascular}'' & ``\N{caseStateNodules}'' \\[5pt]
Decision-first & ``\N{caseOursVascular}'' & ``\N{caseOursNodules}'' \\
\bottomrule
\end{tabular}
\par\vspace{4pt}
\begin{minipage}{\linewidth}\footnotesize
Excerpts preserve the original wording; unrelated report sentences are omitted.
\end{minipage}
\endgroup
\end{table}

%% file: tables/intervention-cases.tex
% CONTENT CONTRACT (SPINE C5/C7): three distinct paired CT cases.
% Rows: input masking (hernia); imposed positive (nodule); imposed negative (artery).
% Columns: intervention/finding/case ID; original output; intervened output.
% Case identifiers and raw report excerpts use occCase*/flip* bindings.
% The empty hernia extraction means the finding is omitted, not an empty report.
% CT is held fixed only for the two decision-edit rows. No cohort score is repeated.
\begin{table}[!htbp]
\centering\small
\caption{Paired report excerpts under input and decision interventions. The first row changes the CT input; the other rows edit one generated answer while holding the CT fixed. Italic text describes the absence of a finding statement. Quoted text is reproduced from the stored outputs.}
\label{tab:intervention-cases}
\setlength{\tabcolsep}{4pt}
\renewcommand{\arraystretch}{1.15}
\begin{tabular}{@{}L{0.24\linewidth}L{0.35\linewidth}L{0.35\linewidth}@{}}
\toprule
Intervention / finding & Before & After \\
\midrule
\textbf{Zero input portion 2}\newline
Hiatal hernia\newline
\texttt{\N{occCaseVolume}}
& \textbf{Decision: yes}\newline ``\N{occCaseQuote}''
& \textbf{Decision: no}\newline \emph{No hiatal-hernia statement.} \\[5pt]
\textbf{Impose yes}\newline
Lung nodule\newline
\texttt{\N{flipAbsentVolume}}\newline
Reference label: absent
& \textbf{Decision: no}\newline ``\N{flipAbsentBefore}''
& \textbf{Edited answer: yes}\newline ``\N{flipAbsentAfter}'' \\[5pt]
\textbf{Impose no}\newline
Arterial wall calcification\newline
\texttt{\N{flipSiblingVolume}}
& \textbf{Decision: yes}\newline ``\N{flipSiblingBefore}''
& \textbf{Edited answer: no}\newline ``\N{flipSiblingAfter}'' \\
\bottomrule
\end{tabular}
\end{table}

%% file: figures/occlusion.tex
% Percentages are formatted once from full-precision loss rates in occlusion.json.
% Each literal N key is intentionally present for the manuscript's number scanner.
\begin{figure}[!htbp]
\centering
\begingroup
\definecolor{occlusionBlue}{RGB}{35,98,145}
\newcommand{\occHeatCell}[3]{%
  \pgfmathsetmacro{\occHeatShade}{min(100,2*(#3))}%
  \pgfmathtruncatemacro{\occHeatDark}{(#3)>30}%
  \ifnum\occHeatDark=1\def\occHeatInk{white}\else\def\occHeatInk{black}\fi
  \fill[occlusionBlue!\occHeatShade!white] (#1-0.45,-#2-0.43) rectangle (#1+0.45,-#2+0.43);
  \node[text=\occHeatInk,font=\scriptsize] at (#1,-#2)
    {#3};
}
\begin{tikzpicture}[x=0.82cm,y=0.43cm,font=\scriptsize,inner sep=1pt]
\node[anchor=east,font=\scriptsize\bfseries] at (-0.85,0) {Finding};
\node[anchor=east,font=\scriptsize\bfseries] at (0.2,0) {$n$};
\node[font=\scriptsize\bfseries] at (3.5,1.95) {Masked portion of the processed frame sequence};
\node[anchor=west] at (0.55,1.15) {Caudal};
\node[anchor=east] at (6.45,1.15) {Cranial};
\draw[->,thin] (1.65,1.15)--(5.15,1.15);
\foreach \col in {1,...,6} {\node[font=\scriptsize\bfseries] at (\col,0) {\col};}
\node[align=center,font=\scriptsize\bfseries] at (7.6,0.35) {Unchanged\\repeat};
\draw[black!30,thin] (6.85,0.8)--(6.85,-18.5);
\node[anchor=east] at (-0.85,-1) {Medical material};
\node[anchor=east,font=\scriptsize\ttfamily] at (0.2,-1) {\N{occHeatN00}};
\occHeatCell{1}{1}{\N{occHeat00P1}}
\occHeatCell{2}{1}{\N{occHeat00P2}}
\occHeatCell{3}{1}{\N{occHeat00P3}}
\occHeatCell{4}{1}{\N{occHeat00P4}}
\occHeatCell{5}{1}{\N{occHeat00P5}}
\occHeatCell{6}{1}{\N{occHeat00P6}}
\occHeatCell{7.6}{1}{\N{occHeatR00}}

\node[anchor=east] at (-0.85,-2) {Arterial wall calcification};
\node[anchor=east,font=\scriptsize\ttfamily] at (0.2,-2) {\N{occHeatN01}};
\occHeatCell{1}{2}{\N{occHeat01P1}}
\occHeatCell{2}{2}{\N{occHeat01P2}}
\occHeatCell{3}{2}{\N{occHeat01P3}}
\occHeatCell{4}{2}{\N{occHeat01P4}}
\occHeatCell{5}{2}{\N{occHeat01P5}}
\occHeatCell{6}{2}{\N{occHeat01P6}}
\occHeatCell{7.6}{2}{\N{occHeatR01}}

\node[anchor=east] at (-0.85,-3) {Cardiomegaly};
\node[anchor=east,font=\scriptsize\ttfamily] at (0.2,-3) {\N{occHeatN02}};
\occHeatCell{1}{3}{\N{occHeat02P1}}
\occHeatCell{2}{3}{\N{occHeat02P2}}
\occHeatCell{3}{3}{\N{occHeat02P3}}
\occHeatCell{4}{3}{\N{occHeat02P4}}
\occHeatCell{5}{3}{\N{occHeat02P5}}
\occHeatCell{6}{3}{\N{occHeat02P6}}
\occHeatCell{7.6}{3}{\N{occHeatR02}}

\node[anchor=east] at (-0.85,-4) {Pericardial effusion};
\node[anchor=east,font=\scriptsize\ttfamily] at (0.2,-4) {\N{occHeatN03}};
\occHeatCell{1}{4}{\N{occHeat03P1}}
\occHeatCell{2}{4}{\N{occHeat03P2}}
\occHeatCell{3}{4}{\N{occHeat03P3}}
\occHeatCell{4}{4}{\N{occHeat03P4}}
\occHeatCell{5}{4}{\N{occHeat03P5}}
\occHeatCell{6}{4}{\N{occHeat03P6}}
\occHeatCell{7.6}{4}{\N{occHeatR03}}

\node[anchor=east] at (-0.85,-5) {Coronary artery wall calcification};
\node[anchor=east,font=\scriptsize\ttfamily] at (0.2,-5) {\N{occHeatN04}};
\occHeatCell{1}{5}{\N{occHeat04P1}}
\occHeatCell{2}{5}{\N{occHeat04P2}}
\occHeatCell{3}{5}{\N{occHeat04P3}}
\occHeatCell{4}{5}{\N{occHeat04P4}}
\occHeatCell{5}{5}{\N{occHeat04P5}}
\occHeatCell{6}{5}{\N{occHeat04P6}}
\occHeatCell{7.6}{5}{\N{occHeatR04}}

\node[anchor=east] at (-0.85,-6) {Hiatal hernia};
\node[anchor=east,font=\scriptsize\ttfamily] at (0.2,-6) {\N{occHeatN05}};
\occHeatCell{1}{6}{\N{occHeat05P1}}
\occHeatCell{2}{6}{\N{occHeat05P2}}
\occHeatCell{3}{6}{\N{occHeat05P3}}
\occHeatCell{4}{6}{\N{occHeat05P4}}
\occHeatCell{5}{6}{\N{occHeat05P5}}
\occHeatCell{6}{6}{\N{occHeat05P6}}
\occHeatCell{7.6}{6}{\N{occHeatR05}}

\node[anchor=east] at (-0.85,-7) {Lymphadenopathy};
\node[anchor=east,font=\scriptsize\ttfamily] at (0.2,-7) {\N{occHeatN06}};
\occHeatCell{1}{7}{\N{occHeat06P1}}
\occHeatCell{2}{7}{\N{occHeat06P2}}
\occHeatCell{3}{7}{\N{occHeat06P3}}
\occHeatCell{4}{7}{\N{occHeat06P4}}
\occHeatCell{5}{7}{\N{occHeat06P5}}
\occHeatCell{6}{7}{\N{occHeat06P6}}
\occHeatCell{7.6}{7}{\N{occHeatR06}}

\node[anchor=east] at (-0.85,-8) {Emphysema};
\node[anchor=east,font=\scriptsize\ttfamily] at (0.2,-8) {\N{occHeatN07}};
\occHeatCell{1}{8}{\N{occHeat07P1}}
\occHeatCell{2}{8}{\N{occHeat07P2}}
\occHeatCell{3}{8}{\N{occHeat07P3}}
\occHeatCell{4}{8}{\N{occHeat07P4}}
\occHeatCell{5}{8}{\N{occHeat07P5}}
\occHeatCell{6}{8}{\N{occHeat07P6}}
\occHeatCell{7.6}{8}{\N{occHeatR07}}

\node[anchor=east] at (-0.85,-9) {Atelectasis};
\node[anchor=east,font=\scriptsize\ttfamily] at (0.2,-9) {\N{occHeatN08}};
\occHeatCell{1}{9}{\N{occHeat08P1}}
\occHeatCell{2}{9}{\N{occHeat08P2}}
\occHeatCell{3}{9}{\N{occHeat08P3}}
\occHeatCell{4}{9}{\N{occHeat08P4}}
\occHeatCell{5}{9}{\N{occHeat08P5}}
\occHeatCell{6}{9}{\N{occHeat08P6}}
\occHeatCell{7.6}{9}{\N{occHeatR08}}

\node[anchor=east] at (-0.85,-10) {Lung nodule};
\node[anchor=east,font=\scriptsize\ttfamily] at (0.2,-10) {\N{occHeatN09}};
\occHeatCell{1}{10}{\N{occHeat09P1}}
\occHeatCell{2}{10}{\N{occHeat09P2}}
\occHeatCell{3}{10}{\N{occHeat09P3}}
\occHeatCell{4}{10}{\N{occHeat09P4}}
\occHeatCell{5}{10}{\N{occHeat09P5}}
\occHeatCell{6}{10}{\N{occHeat09P6}}
\occHeatCell{7.6}{10}{\N{occHeatR09}}

\node[anchor=east] at (-0.85,-11) {Lung opacity};
\node[anchor=east,font=\scriptsize\ttfamily] at (0.2,-11) {\N{occHeatN10}};
\occHeatCell{1}{11}{\N{occHeat10P1}}
\occHeatCell{2}{11}{\N{occHeat10P2}}
\occHeatCell{3}{11}{\N{occHeat10P3}}
\occHeatCell{4}{11}{\N{occHeat10P4}}
\occHeatCell{5}{11}{\N{occHeat10P5}}
\occHeatCell{6}{11}{\N{occHeat10P6}}
\occHeatCell{7.6}{11}{\N{occHeatR10}}

\node[anchor=east] at (-0.85,-12) {Pulmonary fibrotic sequela};
\node[anchor=east,font=\scriptsize\ttfamily] at (0.2,-12) {\N{occHeatN11}};
\occHeatCell{1}{12}{\N{occHeat11P1}}
\occHeatCell{2}{12}{\N{occHeat11P2}}
\occHeatCell{3}{12}{\N{occHeat11P3}}
\occHeatCell{4}{12}{\N{occHeat11P4}}
\occHeatCell{5}{12}{\N{occHeat11P5}}
\occHeatCell{6}{12}{\N{occHeat11P6}}
\occHeatCell{7.6}{12}{\N{occHeatR11}}

\node[anchor=east] at (-0.85,-13) {Pleural effusion};
\node[anchor=east,font=\scriptsize\ttfamily] at (0.2,-13) {\N{occHeatN12}};
\occHeatCell{1}{13}{\N{occHeat12P1}}
\occHeatCell{2}{13}{\N{occHeat12P2}}
\occHeatCell{3}{13}{\N{occHeat12P3}}
\occHeatCell{4}{13}{\N{occHeat12P4}}
\occHeatCell{5}{13}{\N{occHeat12P5}}
\occHeatCell{6}{13}{\N{occHeat12P6}}
\occHeatCell{7.6}{13}{\N{occHeatR12}}

\node[anchor=east] at (-0.85,-14) {Mosaic attenuation pattern};
\node[anchor=east,font=\scriptsize\ttfamily] at (0.2,-14) {\N{occHeatN13}};
\occHeatCell{1}{14}{\N{occHeat13P1}}
\occHeatCell{2}{14}{\N{occHeat13P2}}
\occHeatCell{3}{14}{\N{occHeat13P3}}
\occHeatCell{4}{14}{\N{occHeat13P4}}
\occHeatCell{5}{14}{\N{occHeat13P5}}
\occHeatCell{6}{14}{\N{occHeat13P6}}
\occHeatCell{7.6}{14}{\N{occHeatR13}}

\node[anchor=east] at (-0.85,-15) {Peribronchial thickening};
\node[anchor=east,font=\scriptsize\ttfamily] at (0.2,-15) {\N{occHeatN14}};
\occHeatCell{1}{15}{\N{occHeat14P1}}
\occHeatCell{2}{15}{\N{occHeat14P2}}
\occHeatCell{3}{15}{\N{occHeat14P3}}
\occHeatCell{4}{15}{\N{occHeat14P4}}
\occHeatCell{5}{15}{\N{occHeat14P5}}
\occHeatCell{6}{15}{\N{occHeat14P6}}
\occHeatCell{7.6}{15}{\N{occHeatR14}}

\node[anchor=east] at (-0.85,-16) {Consolidation};
\node[anchor=east,font=\scriptsize\ttfamily] at (0.2,-16) {\N{occHeatN15}};
\occHeatCell{1}{16}{\N{occHeat15P1}}
\occHeatCell{2}{16}{\N{occHeat15P2}}
\occHeatCell{3}{16}{\N{occHeat15P3}}
\occHeatCell{4}{16}{\N{occHeat15P4}}
\occHeatCell{5}{16}{\N{occHeat15P5}}
\occHeatCell{6}{16}{\N{occHeat15P6}}
\occHeatCell{7.6}{16}{\N{occHeatR15}}

\node[anchor=east] at (-0.85,-17) {Bronchiectasis};
\node[anchor=east,font=\scriptsize\ttfamily] at (0.2,-17) {\N{occHeatN16}};
\occHeatCell{1}{17}{\N{occHeat16P1}}
\occHeatCell{2}{17}{\N{occHeat16P2}}
\occHeatCell{3}{17}{\N{occHeat16P3}}
\occHeatCell{4}{17}{\N{occHeat16P4}}
\occHeatCell{5}{17}{\N{occHeat16P5}}
\occHeatCell{6}{17}{\N{occHeat16P6}}
\occHeatCell{7.6}{17}{\N{occHeatR16}}

\node[anchor=east] at (-0.85,-18) {Interlobular septal thickening};
\node[anchor=east,font=\scriptsize\ttfamily] at (0.2,-18) {\N{occHeatN17}};
\occHeatCell{1}{18}{\N{occHeat17P1}}
\occHeatCell{2}{18}{\N{occHeat17P2}}
\occHeatCell{3}{18}{\N{occHeat17P3}}
\occHeatCell{4}{18}{\N{occHeat17P4}}
\occHeatCell{5}{18}{\N{occHeat17P5}}
\occHeatCell{6}{18}{\N{occHeat17P6}}
\occHeatCell{7.6}{18}{\N{occHeatR17}}

\node[anchor=west] at (0.55,-19.35) {Report-label loss (\%)};
\foreach \value in {0,10,...,50} {
  \pgfmathsetmacro{\occLegendX}{0.55+0.9*\value/10}
  \pgfmathsetmacro{\occLegendShade}{2*\value}
  \fill[occlusionBlue!\occLegendShade!white,draw=black!15,line width=0.2pt]
    (\occLegendX,-20.65) rectangle (\occLegendX+0.9,-19.95);
  \node at (\occLegendX+0.45,-21.15) {\value};
}
\end{tikzpicture}
\endgroup
\caption{Input masking and loss of reference-positive report labels.
All \N{evalVolumes} corrected validation volumes are evaluated under each mask and an unchanged repeat.
For each finding, $n$ counts volumes with a positive reference label and a positive label extracted from the unmodified report.
Cells show the percentage of these reports scored negative after the indicated intervention; the repeat column uses an unchanged input.
Each mask sets one sixth of the concatenated, processed axial-frame sequence to zero, including any repeated padding frames; indices and sequence length are preserved.
Portions run from the caudal to the cranial end of the displayed sequence, not equal physical anatomical regions.
}
\label{fig:occlusion}
\end{figure}